\pdfoutput=1

\documentclass[11pt]{article}

\usepackage[]{acl}

\usepackage{times}
\usepackage{latexsym}
\usepackage{amsmath}

\usepackage{multirow}
\usepackage{booktabs}
\usepackage{graphicx}
\usepackage{subcaption}
\usepackage{enumitem}
\usepackage{float}
\usepackage{array}
\usepackage[]{xcolor}

\usepackage[T1]{fontenc}

\usepackage[utf8]{inputenc}

\usepackage{microtype}

\title{Infusing Emotions into Task-oriented Dialogue Systems: Understanding, Management, and Generation}

\author{Shutong Feng$^{*}$, Hsien-chin Lin$^{*}$, Christian Geishauser$^{*}$, Nurul Lubis, Carel van Niekerk,\\{\bf Michael Heck, Benjamin Ruppik, Renato Vukovic, Milica Ga\v{s}i\'{c}} \\
  Heinrich Heine University Düsseldorf, Germany\\
  \texttt{\small{\{fengs,linh,geishaus,lubis,niekerk,heckmi,ruppik,revuk100,gasic\}@hhu.de}} \\}

\begin{document}
\maketitle

\begin{abstract}

Emotions are indispensable in human communication, but are often overlooked in task-oriented dialogue (ToD) modelling, where the task success is the primary focus. While existing works have explored user emotions or similar concepts in some ToD tasks, none has so far included emotion modelling into a fully-fledged ToD system nor conducted interaction with human or simulated users. In this work, we incorporate emotion into the complete ToD processing loop, involving understanding, management, and generation. To this end, we extend the EmoWOZ dataset \citep{feng-etal-2022-emowoz} with system affective behaviour labels. Through interactive experimentation involving both simulated and human users, we demonstrate that our proposed framework significantly enhances the user's emotional experience as well as the task success.


\end{abstract}

\section{Introduction}
\def\thefootnote{*}\footnotetext{These authors contributed equally to this work.} 

In recent years, conversational artificial intelligence (AI) has become increasingly prevalent in various domains, providing users with interactive and personalised experiences. Emotions play a crucial role in human communication and can influence the way individuals perceive, process, and react to information \citep{Ekman1992AnAF}. Consequently, incorporating emotions into conversational AI has emerged as a promising avenue for improving user experience and creating more human-like interactions \citep{Picard2000AffectiveC}. 

Task-oriented dialogue (ToD) systems, an important genre of conversational AI, are designed to assist users in fulfilling specific tasks or queries. In contrast to chit-chat or open-domain dialogue systems, which focus on creating engaging and entertaining conversations, ToD systems interact with users in a more structured way with a clear objective under specific domains \citep{Jurafsky2009}. While significant advancements have been made in natural language processing and ToD systems, there remains a critical challenge in creating systems that can understand and respond to not only the informational needs of users but also their emotional states.

In ToD, emotion is centred around the user goal, making it more contextual and subtle \cite{feng-etal-2022-emowoz}. A recent study has shown that the valence of user emotion in ToD positively correlates with dialogue success \cite{emous}. This observation aligns with a number of emotional theories. For example, the appraisal theory of emotion argues that emotion is the result of our evaluation of a situation \cite{arnold1960emotion, lazarus1966psychological}. In relation to a ToD user goal, it is straightforward to see how task fulfilment would lead to positive emotions and failures to negative ones. Similarly, the Ortony-Clore-Collins (OCC) model of emotion states that emotion is the result of elicitation by events, agents, and objects \cite{ortony_clore_collins_1988}. \citet{feng-etal-2022-emowoz} have drawn the connection between the OCC model and user emotions in ToD. Therefore, besides inferring emotional states from dialogue utterances, an agent also needs to reason about emotion-generating situations and to utilise this information to achieve dialogue success.

The integration of emotion into the full ToD pipeline has been a long-standing interest \citep{affect-pomdp,tfsm}. Yet, early works explored analytical solutions in constrained set-ups, which hindered their applications in more complicated scenarios. Recently, a number of resources emerged for studying user affect in ToDs, e.g. emotion, sentiment, or satisfaction \citep{maia-dqe,feng-etal-2022-emowoz}. This has motivated efforts to model user emotion via data-driven approaches, such as emotional user simulation~\citep{emous} and user emotion recognition~\citep{feng-etal-2023-chatter,emotod}. However, to the best of our knowledge, no work so far has combined these emotional aspects into a fully-fledged dialogue system and an interactive pipeline where emotions play a role in understanding, generation, as well as management of the conversation.

To achieve this, we need to endow the dialogue system with the ability to respond with an affective behaviour, closing the emotional loop between the user and the system in ToDs. Towards this goal, we make the following contributions:


\begin{itemize}
\setlength{\itemindent}{0em}
\setlength\itemsep{0em}
    \item We extend EmoWOZ, a large-scale ToD dataset for user emotions \citep{feng-etal-2022-emowoz}, with annotations for \emph{affective conduct} in 71k system utterances. To the best of our knowledge, this is the first large-scale and open-source corpus dedicated to the system's affective behaviour in ToDs.
    \item We incorporate emotion in the complete ToD interaction loop for understanding, management, and generation by building a modular system around an emotion-\emph{aware} and emotion-\emph{expressive} policy. We also build an emotional LLM-based end-to-end ToD system that involves emotion in understanding and generation.
    \item For our modular system, we train our dialogue policy via reinforcement learning (RL) on the natural language level, leveraging emotions and task success as reward signals. We train the end-to-end system on our newly collected dataset via supervised learning (SL). For both systems, we show through interactive evaluation
    that emotion in the ToD loop can enhance user's emotional experience as well as the task success. This highlights the importance of modelling emotions in ToDs.
\end{itemize}

\begin{figure*}[h]
    \centering
        \begin{subfigure}[b]{0.47\textwidth}   
            \centering 
            \includegraphics[width=\textwidth]{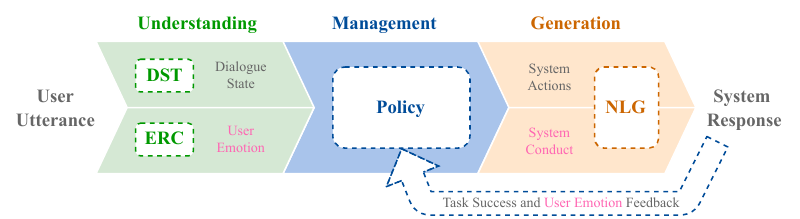}
            \caption{\small Modular Systems}   
            \label{fig:new-rl-pipeline}
        \end{subfigure}
        \hfill
        \begin{subfigure}[b]{0.49\textwidth}   
            \centering 
            \includegraphics[width=\textwidth]{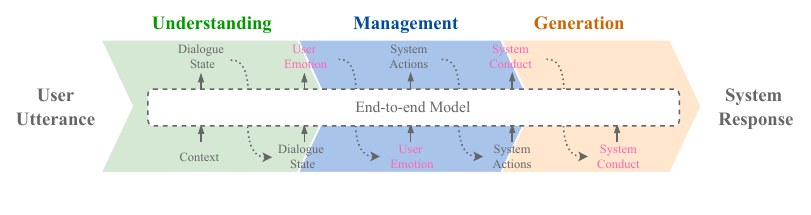}
            \caption{\small End-to-end Systems}    
            \label{fig:emotion-in-e2e}
        \end{subfigure}
    \label{fig:emoloop-complete}
    \caption{Infusing emotions into modular and end-to-end ToD systems.}
\end{figure*}


\section{Related Work}

In this section, we discuss related works on incorporating emotion in each stage of ToD pipeline: understanding, management, and generation. These stages are modelled explicitly with multiple models in modular systems and implicitly with a unified model in end-to-end systems \citep{simpletod,emotod}.

\subsection{Understanding User Emotion}
Modular ToD systems rely on natural language understanding (NLU) and dialogue state tracking (DST) modules to translate and accumulate semantic concepts related to user goals. 
Typically, these semantic concepts are strictly limited to those defined in the ontology, i.e. domains, slots, and values the system can talk about. 

Given its potential as an important piece of information for the system's subsequent decision-making, emotion can be considered as part of the dialogue state. \citet{feng-etal-2022-emowoz} showed that multi-task training a DST model for emotion recognition simultaneously improves its joint goal accuracy, suggesting the complementarity between DST and emotion recognition in conversation (ERC). Recently, \citet{emotod} modelled user emotion as an intermediate task in end-to-end ToD systems and improved overall system performance. 
Standalone ERC models dedicated to ToDs \citep{unisa,feng2023affect} can be used in modular systems in parallel with any DST to extend the dialogue state with user emotions.

\subsection{Dialogue Management with Emotion Feedback}
\label{sec:background-emotion-aware-policy}
In ToD, one way to train the dialogue policy is via RL to maximise task success, indicated at the end of the dialogue based on user goal fulfilment \cite{Levin97astochastic, Kwan2023-surveydialoguepolicies}. Since user emotion is highly associated with task success \cite{emous}, it is intuitive to leverage user emotion during the dialogue for providing more dense and diverse reward signals. \citet{affect-pomdp} incorporate user emotion into the policy state by modelling affective dialogue management through a factored partially-observable Markov decision process (POMDP) and analytically find an optimal policy.
This is however neither feasible for larger problems, nor has this been integrated in interactive set-ups. \citet{ZHANG2021122} addressed the delayed reward problem in dialogue policy learning with a predefined emotion-based turn-level reward. 
\citet{esdp} consider the difference between the user's positive emotion intensity and the next turn's emotion utility value for top-k action selection. 
We take a step further by incorporating emotion in policy state \textit{and} reward function. We then leverage emotion in RL to find optimal semantic actions \textit{and} affective expression of the system, which has not been explored before. 


\subsection{Generating Affective Response}
The natural language generation (NLG) module in ToD systems realises semantic actions from the policy into natural language. Traditionally, ToD NLG focuses on translating task-related semantic actions and overlooks other aspects of system responses such as emotion. There have been efforts to create datasets which help enrich ToD system responses with chit-chat \citep{sun-etal-2021-adding,chen-etal-2022-ketod,fusechat}. \citet{emotod} attempted to refine end-to-end system output with a large language model (LLM) under a chain-of-thought framework to enhance simulated system empathy. 
Different from previous works, we aim to enrich system response with the subtle affective conduct jointly with dialogue actions in a fully controllable approach.

\subsection{Simulating User Emotional Behaviour}
User simulators (USs) simulate user behaviour in ToDs. Although they are not a part of the system, they play essential roles in training dialogue policy via RL and serving as an efficient evaluation platform for dialogue policy \citep{658991}. Most existing USs focus on modelling user's behaviour in terms of semantic actions and natural language by taking system semantic actions \citep{kreyssig-etal-2018-neural,lin-etal-2021-domain,lin-etal-2022-gentus}. \citet{ZHANG2021122} built a US that additionally incorporated handcrafted emotion transitions in different situations. \citet{satactutt} used a data-driven approach and simulated satisfaction levels along with the intent and the utterance. \citet{emous} further proposed data-driven EmoUS to model more nuanced user emotions with enhanced controllability via user persona settings. This motivates us to move one step further to capture more fine-grained affective expressions of the system from natural language response directly. 

\section{EmoWOZ 2.0: A Fully Emotion-annotated ToD Dataset}
To study emotion in real-world interactions between users and human operators in the ToD setting,
we extend EmoWOZ \citep{feng-etal-2022-emowoz} 
by further annotating the \emph{affective behaviour} of the system, which is acted by human-beings. We call this dataset with extended labels \emph{EmoWOZ 2.0}.~\footnote{EmoWOZ 2.0 is released under CC By 4.0 NC license, following the original EmoWOZ release. The dataset can be found at \url{https://gitlab.cs.uni-duesseldorf.de/general/dsml/emowoz-2.0-public/}}

In ToDs, the user and the system play different roles. Users may express a wide range of emotions during interactions based on their goals and experiences with the operator. The system is responsible for managing and facilitating the conversation and is supposed to behave professionally and politely to achieve the goal. Therefore, it is necessary to consider different sets of affective behaviours in the user and the system respectively. We refer to the concept of the operator's affective behaviour as \textbf{affective conduct}, or \textbf{conduct} for short.

\paragraph{Annotation Scheme}




According to studies on customer satisfaction in business \citep{doi:10.1177/1094670511410304}, competent operators in ToD try to guide user emotion towards positive valence by making use of subtle emotion in their response while providing correct information. By considering the set of user emotions in EmoWOZ and the OCC emotion model (detailed justification in Appendix \ref{sec:occ-justification}), we arrive at five affective conduct classes: 
\begin{itemize}[itemsep=0pt,parsep=0pt,left=0pt]
    \item \textbf{Neutral:} the operator does not explicitly make use of any affective conduct.
    \item \textbf{Compassionate:} the operator is sympathetic about user's situation, usually in response to a fearful/disappointed user in an unpleasant situation.
    \item \textbf{Apologetic:} the operator apologises for their mistake, usually in response to a dissatisfied user.
    \item \textbf{Enthusiastic:} the operator is feeling happy for the user or showing extra eagerness to help. This conduct takes place usually in response to a neutral or excited user.
    \item \textbf{Appreciative:} the operator acknowledges the -- at least partial -- task success, usually signalled as user's satisfaction.
\end{itemize}



\paragraph{Annotation Set-up}
We annotated the conduct for all operator utterances in the MultiWOZ subset of EmoWOZ. Machine-generated system responses in the DialMAGE subset came from a template NLG, which we considered to have neutral conduct because those templates aimed to express actions concisely rather than conveying emotions by design.

We followed the data collection and quality assurance set-up of EmoWOZ and conducted the annotation via the Amazon Mechanical Turk platform. 
Details and an illustration of the annotation interface can be found in Appendix \ref{sec:emowoz2.0-interface}.

\paragraph{Annotation Quality} Each utterance has been annotated by at least three annotators. The inter-annotator agreement as measured with Fleiss' Kappa is 0.647, suggesting substantial inter-annotator agreement. The annotator confusion matrix and label distribution can be found in Appendix \ref{sec:emowoz2.0-confusion} and \ref{sec:emowoz2.0-statistics}, respectively. 

\section{Infusing Emotions into ToD Systems}
\label{sec:modules}
We propose to incorporate emotion into the full interactive ToD pipeline, which is primarily comprised of three stages: understanding, management, and generation. We aim for understanding to accurately recognise the user's emotion in addition to the task-centred dialogue state. For dialogue management, we make use of emotion for optimal action selection. 
Lastly, we additionally condition the natural language generation on the system conduct to generate more diverse and emotion-aware responses. These are realised in each modular system component individually (Section \ref{sec:emotion-in-understanding} to \ref{sec:emotion-in-generation}) and as intermediate tasks in the unified model in end-to-end systems (Section \ref{sec:e2e-llama}).
\footnote{The code of pipeline systems, end-to-end systems, and the user simulator can be found at \url{https://gitlab.cs.uni-duesseldorf.de/general/dsml/emoloop-public/}}


\subsection{Expanding Dialogue State with Emotion}
\label{sec:emotion-in-understanding}
In our modular system, we use an ERC model in parallel with a DST model. This allows a flexible selection of DST and the associated ontology. The inferred user emotion is appended to the dialogue state.

For ERC, we use the ContextBERT-ERToD model \citep{feng-etal-2023-chatter} as our user emotion recognition front-end because of its good ERC ability and fast inference. It is a BERT-based classification model \citep{devlin-etal-2019-bert} that considers dialogue context and state in addition to the user utterance. It reports a weighted F1 score of 83.9\% for emotions excluding neutral.

For DST, we use the SetSUMBT model \citep{van-niekerk-etal-2021-uncertainty}. This model, based on the RoBERTa language model \citep{liu2019roberta} and a recurrent context tracker adopts a picklist approach to DST. 
Specifically, we employ the \texttt{Ensemble-} \texttt{Distribution-Distilled} variant of SetSUMBT, a refined version that distils knowledge from an ensemble of models. This version reports a joint goal accuracy of 51.22\% on MultiWOZ. The architectural design of SetSUBMT also allows transferability to new domains, and such an ability has been exemplified with a transformer-based dialogue policy under a continual learning set-up \citep{10491378}.

\subsection{Emotion-aware Dialogue Policy}
\label{sec:emo-ddpt}

For dialogue management in the modular system, we build a dialogue policy that considers the user emotion in the input and produces an emotion-augmented system output. We utilise the dynamic dialogue policy transformer (DDPT) architecture \cite{geishauser2022-ddpt} since it was built for optimising dialogue policies that require extendable input and output, which facilitate the adaptation to new domains and ontologies. The dialogue policy leverages emotions in three ways: considering user emotion in the input, generating system affective conduct in the output, and considering user emotion in the reward for RL.

\paragraph{Emotion Input and Output}
The user emotion, as a part of the dialogue state, is incorporated into the dialogue state through embedding the perceived user emotion with RoBERTa. For semantic action selection, DDPT produces a sequence of domain-intent-slot triplets auto-regressively through its transformer decoder, e.g. \texttt{restaurant-inform-phone, restaurant-request-food}, until a stop token is generated. In order to predict \emph{emotional} system conduct, after DDPT outputs the semantic actions, 
we decode the sequence for one more step to generate the system conduct action, considering the perceived user emotion from the dialogue state.

\paragraph{Emotion Augmented Reward}

We incorporate user emotion into the reward for RL by considering the associated sentiment. More specifically, we define $c(\mathrm{satisfied})=1$, $c(\mathrm{dissatisfied})=c(\mathrm{abusive})=-1$, $c(\mathrm{neutral})=0$. For the remaining user emotions that are not elicited by the system, we set $c(\mathrm{emotion})=0$. For any emotion $e$, we multiply $c(e)$ by a hyperparameter $\beta$ to weight the influence of emotion in the reward. Note that utilizing $\beta \cdot c(e)$ directly could encourage the dialogue policy to produce long dialogues with unnecessary turns as long as they produce positive user sentiment. In order to prevent this, we shift $\beta \cdot c(e)$ such that it is at most $0$ by defining the emotion reward for an emotion $e$ as $r_\mathrm{emo}(e) = \beta \cdot c(e) - \beta$.


The emotional reward is combined with the standard reward $r_\mathrm{task}$ in dialogue policy learning that equals $-1$ in every non-terminating turn for encouraging efficiency and either $-T$ or $2T$ for dialogue failure or success, where $T$ denotes the maximum permitted number of turns. The final reward is thus given by $r = r_\mathrm{task} + r_\mathrm{emo}$. We refer to this policy with expanded dialogue state input, expanded dialogue action output, and emotion reward as \textbf{EmoDDPT}.

\subsection{Expressing Emotion in Response}
\label{sec:emotion-in-generation}
Our modular system NLG was built based on the BART model~\cite{lewis2020bart}. We followed existing works to formulate the ToD NLG problem as a sequence-to-sequence task \citep{peng-etal-2020-shot,zhu-etal-2023-convlab} where the input is a sequence containing semantic concepts in textual form (e.g. tuples of [intent, domain, slot, value]), and the output is natural language conveying the semantic meaning. Our model input consists of the user utterance, system semantic actions, and the system conduct. We refer to our system NLG as \textbf{SEC-BART}: a both \textbf{s}emantically and \textbf{e}motionally \textbf{c}onditioned BART. In our ablation study, we used \textbf{SC-BART}, the version that is only conditioned on the semantic actions in the non-emotional ToD pipeline. 

On MultiWOZ, SEC-BART achieves a BLEU score of 34.9 and a slot error rate of 3.6\%, comparable to existing SOTAs \citep{peng-etal-2020-shot,zhu-etal-2023-convlab}. Details of model training and performance can be found in Appendix \ref{sec:nlg-appendix}.

\subsection{Emotional End-to-end System}
\label{sec:e2e-llama}

We follow the work of \citet{emotod}, where ERC is added as an intermediate task in the end-to-end ToD modelling, i.e. emotion is incorporated in the understanding stage. We further consider emotion in the generation stage by predicting the system conduct in the end-to-end pipeline, as illustrated in Figure \ref{fig:emotion-in-e2e}. To this end, we build a LLaMA-based end-to-end ToD system that involves emotion in both understanding and generation, with LLaMA-2-7B \citep{touvron2023llama} as the backbone. As illustrated in Figure \ref{fig:emotion-in-e2e}, it takes dialogue history and the recognised user emotion as input, and then auto-regressively generates the dialogue state, user emotion, system actions, system conduct, and delexicalised natural language response. The response is then lexicalised via database queries based on the intermediately generated dialogue state and system actions. We refer to this end-to-end model as \textbf{EmoLLAMA}. 

We did not train EmoLLAMA via RL with task and emotion feedback from the user simulator because it would take more than 20 days on an A100 40GB to simulate the same number of dialogues as we did to train the EmoDDPT policy in the modular system. We therefore leave efficient training of LLM-based ToD systems via RL as a future research direction.

\subsection{Emotional User Simulation}

Traditionally, user simulators interact with the system on the semantic level for efficiency. To capture more fine-grained expressions of system conducts in natural language, we build \textbf{langEmoUS} based on EmoUS~\citep{emous}. langEmoUS interacts with the system on the natural language level, e.g. it takes the system utterance, user goal, turn information and user persona as inputs and generates user emotion and user utterance.
The turn information represents the dialogue progress, i.e. the turn number. 
Following the setting in ~\citet{emous}, the user persona is extracted from the dialogue history, e.g. if a user is excited to visit a museum in the conversation, then its persona is $\{attraction: excited\}$, when training the user model supervisedly. During inference, the user persona is sampled from the distribution of the corpus.

LangEmoUS achieves macro F1 scores of 0.742 and 0.521 for user sentiment prediction and emotion prediction, respectively, significantly outperforming existing state-of-the-art models \citep{satactutt,emous} (see Appendix~\ref{sec:user-simulator-appendix}).

\section{Experimental Set-up}
\subsection{Modular System Set-up}
\label{sec:models-modular}
\paragraph{EmoLoop} This is our proposed modular system with emotion incorporated for understanding, management, and understanding, as outlined in Figure \ref{fig:new-rl-pipeline} and Figure \ref{fig:policy-training-nl}. It includes the following modules: SetSUMBT DST, ContextBERT-ERToD ERC, $\textrm{EmoDDPT}$ policy, and SEC-BART NLG. EmoDDPT is trained via RL on the natural language level with langEmoUS.

\paragraph{SimpleLoop} This is the non-emotion baseline to EmoLoop. It neither predicts user emotion for the state, uses emotion reward to train the policy, nor generates system conduct for emotional response generation. Specifically, it includes the following modules: SetSUMBT DST, DDPT policy, and SC-BART NLG. DDPT is trained via RL on the natural language level with langEmoUS.

\subsubsection{Dialogue Policy Optimisation}

We implement our system in the ConvLab-3 framework \citep{zhu-etal-2023-convlab}. We pre-trained the policy on MultiWOZ 2.1 \citep{eric-etal-2020-multiwoz}, followed by online RL through interaction with our US.
During RL, in addition to the emotion reward as outlined in Section \ref{sec:emo-ddpt}, we set the task reward as $-1$ in every turn to encourage efficiency, and $80$ or $-40$ for dialogue success or failure. A dialogue is successful if the system provides the requested information to the user and books the correct entities (if possible). For emotional reward, we set $\beta = 2$. We pre-train each policy on MultiWOZ, followed by 15k dialogues with langEmoUS via RL for 6 random seeds. For every 1k dialogues of training, we evaluate the policy for 500 dialogues. We use overall return to select the best checkpoint. All peripheral modules were trained, implemented, and evaluated in the ConvLab-3 environment. 

\paragraph{Language-level RL Training}

\begin{figure}[h]
    \centering
    \includegraphics[width=0.4\textwidth]{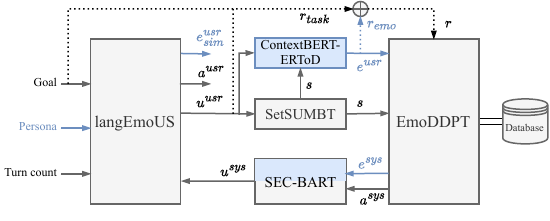}
    \caption{RL training set-up for EmoDDPT.}
    \label{fig:policy-training-nl}
\end{figure}

As illustrated in Figure \ref{fig:policy-training-nl}, our policy, EmoDDPT, interacts with langEmoUS on the natural-language level where the policy actions and conduct ($a^{sys},e^{sys}$) is realised into natural language, $u^sys$ with SEC-BART. The US takes natural-language input and outputs natural-language user utterances $u^{usr}$ after auto-regressively generating the simulation target user emotion $u^{usr}_{sim}$ and user actions $a^{usr}$. The perceived user emotion $e^{usr}$ and dialogue state $s$ are determined by ContextBERT-ERToD and SetSUMBT respectively.



\subsection{End-to-end System Set-up}
\paragraph{EmoLLAMA} This our proposed end-to-end system as described in Section \ref{sec:e2e-llama}.

\paragraph{SimpleLLAMA} This is the non-emotional baseline, which is also used in the work of \citet{emotod}. Compared with EmoLLAMA, it does not consider user emotions as a part of the model input, nor does it auto-regressively predict user emotion and system conduct.


Both EmoLLAMA and SimpleLLAMA are trained and evaluated with EmoWOZ 2.0 using the environment provided by \citet{emotod} and following default parameters. Their interactive evaluations were set up in the ConvLab-3 environment.

\begin{table*}[t]
\centering
\scriptsize
\begin{tabular}{@{}llcccccc@{}}
\toprule

\multirow{2}{*}{\bf System} & \multirow{2}{*}{\bf Type} & \multicolumn{2}{c}{\bf Corpus} & \multicolumn{2}{c}{\bf User Simulator} & \multicolumn{2}{c}{\bf Human} \\
 & & Inform & Success & Success & Sentiment & Success & Sentiment Rating \\ \midrule
SimpleLLAMA & End-to-end & 0.785 & 0.705 & 0.330 & 0.214 & 0.819 & 3.97 \\
EmoLLAMA & End-to-end & \textbf{0.833} & \textbf{0.760} & 0.342 & \textbf{0.250} & \textbf{0.894} & \textbf{4.16} \\ \midrule

SimpleLoop & Modular & 0.700 & 0.621 & 0.556 & 0.337 & 0.798 & 3.85 \\
EmoLoop & Modular & \textbf{0.753} & 0.635 & 0.531 & \textbf{0.405} & \textbf{0.917} & \textbf{4.15} \\ \bottomrule
\end{tabular}
\caption{System evaluation, including corpus-based evaluation, interaction with user simulator and human trial. Values in bold mean best scores with statistically significant difference $p<0.05$.}
\label{tab:evaluation}
\end{table*}

\subsection{Evaluation}
\paragraph{Corpus Evaluation} 
We report \emph{inform} and \emph{success} rates. Inform rate evaluates if the system provides entities from the database that fulfill user’s constraints. Success rate assesses if the system delivers all information requested by the user. To generate each system response, the ground-truth dialogue history was used as system input.  

\paragraph{Interactive Evaluation}
For interactive evaluation, our systems interact with langEmoUS. We report the \emph{success} rate and the average user \emph{sentiment} in simulated dialogues to account for user emotional experience. Specifically, the turn-level sentiment score is $+1$ if the user emotion is positive, $0$ if neutral and $-1$ if negative. User sentiment is determined by the ERC.

\paragraph{Human Trial}
We set up a human trial using the DialCrowd toolkit \citep{huynh-etal-2022-dialcrowd} on the Amazon Mechanical Turk platform. 
We set up two pairs of comparison: 1) SimpleLLAMA vs. EmoLLAMA and 2) SimpleLoop vs. EmoLoop. Volunteers are presented with randomly generated single or multi-domain goals. A goal contains a set of constraints for entities that the user should be looking for (e.g. the price range and the location of a restaurant) and specifies the information they should extract from the system (e.g. the phone number and booking reference of the restaurant). Given a goal, volunteers would need to talk to each system to fulfill the goal. They then give ratings to each of them based on objective (whether the goal has been fulfilled) and subjective metrics (how they feel about the system). Survey questions include objective task success and subjective user sentiment. Details of the website interface and survey questions can be found in Appendix \ref{sec:appendix-human-evaluation}. To obtain more reliable ratings, we filtered out dialogues with poor quality, e.g. containing very short user utterances or non-natural language, and with inconsistent ratings, e.g. system A had better rating in all aspects but overall the rater found system B better. Overall, we collected 203 valid ratings for the SimpleLLAMA-EmoLLAMA comparison and 253 for the SimpleLoop-EmoLoop comparison from 40 unique raters. 


\section{Results and Discussion}
\label{sec:results-and-discussion}
\subsection{Corpus Evaluation}
Although it is not a common practice to evaluate RL-trained modular ToD systems on a corpus, we provide such results for a basic understanding and comparison with end-to-end systems. Our goal is not beating SOTA on task-related metrics, but examining interactive abilities of the system and the role of emotion in it. As shown in Table \ref{tab:evaluation}, incorporating emotion significantly improves inform rate of both types of systems and success rate of the end-to-end system. 

It is not surprising that modular systems underperform when compared with end-to-end systems. Modular systems are trained via RL, which allows the policy to explore more diverse dialogue trajectories but diverges from what a policy can learn from the corpus only. 
This reflects the limitation of corpus evaluation in accounting for ToD system performance, as pointed out by \citet{lubis-etal-2022-dialogue}.



\subsection{Evaluation with User Simulator}
\label{sec:evaluation-us}


In interactive evaluation, both EmoLoop and EmoLLAMA perform significantly better in terms of average sentiment than their respective non-emotional baseline while maintaining the same level of success rate. For end-to-end models, despite the fact that they are not optimised via RL with the simulated user, the average sentiment in the simulated user also improves significantly. 

When comparing performance across system types, modular systems perform better than end-to-end models on task success and simulated user sentiment since modular system policies have been optimised for the simulated user via RL. SimpleLLAMA and EmoLLAMA, trained via SL only, cannot adequately cope with the more diverse user goals and situations of the simulated user. This motivates our future work to leverage the simulated user and to train end-to-end systems via RL.



\begin{figure}[h]
    \centering
    \includegraphics[width=0.4\textwidth]{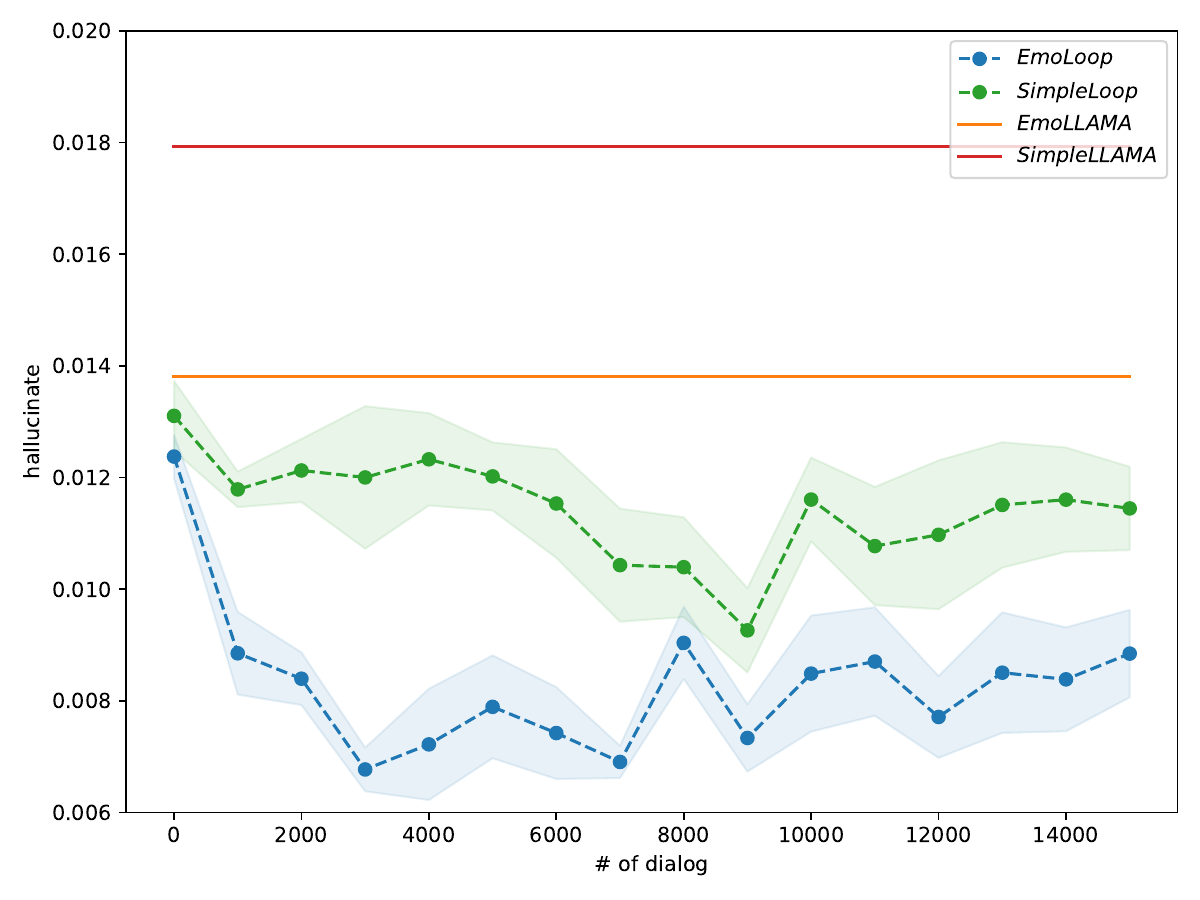}
    \caption{The average hallucination rate of modular systems during RL training with langEmoUS. For end-to-end systems, we report hallucination rate after SL.}
    \label{fig:hallucinate}
\end{figure}

\begin{figure}[h]
    \centering
    \includegraphics[width=0.4\textwidth]{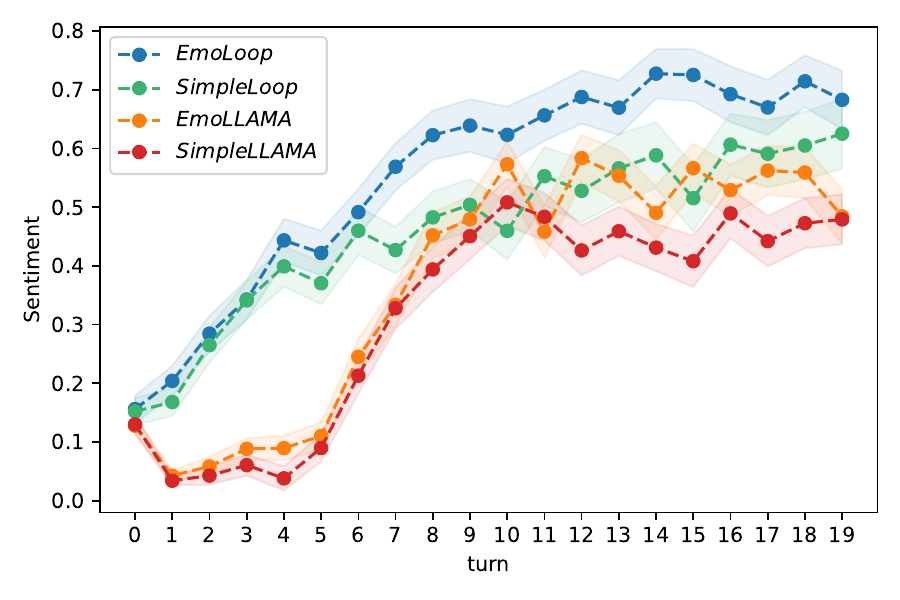}
    \caption{Average sentiment at different turn positions during language-level interaction with langEmoUS. }
    \label{fig:sentiment-progression}
\end{figure}

\paragraph{Hallucination} In ToD, a hallucination is defined as a value in the system response that is not supposed to be informed according to system actions. As shown in Figure~\ref{fig:hallucinate}, the hallucination rate of each type of systems is improved as emotion is incorporated into the pipeline. The hallucination rate is lowered from 1.8\% for SimpleLLAMA to 1.4\% for EmoLLAMA. We observe that end-to-end systems are more prone to the hallucination problem than modular systems as slot placeholders in the delexicalised end-to-end system response do not always match the intermediately generated dialogue actions. 
Hallucination rates of SimpleLoop and EmoLoop are around 1.3\% at the beginning of the interactive RL training and continue to improve as the RL progresses.


\paragraph{Progression of User Sentiment in Dialogues}

Figure \ref{fig:sentiment-progression} shows the average sentiment of langEmoUS at each turn of interactions with our systems. The sentiment level of langEmoUS becomes more positive as the dialogue progresses and moves towards user goal completion in all systems. The primary difference between modular systems and end-to-end systems is that in earlier turns, modular systems are able to satisfy the simulated user better, as illustrated in higher and more positive sentiment level before turn 8.

\subsection{Human Trials}
We carried out human trials to compare two pairs of systems in Table \ref{tab:evaluation}. Within each pair of comparison, the emotion-incorporating model significantly outperforms its non-emotion version in terms of both the success rate and user sentiment. This further confirms our findings from corpus and user simulator evaluations.
Example dialogue excerpts are given in Appendix \ref{sec:appendix-human-evaluation-example} to exemplify how emotional ToD systems made use of affective conduct in case of neutral and unsuccessful interactions.

Although human ratings across system types are not directly comparable, it is noteworthy that the absolute improvement from SimpleLLAMA to EmoLLAMA ($\Delta \textrm{Success}=0.075$, $\Delta \textrm{Sentiment}=0.19$) is smaller than that from SimpleLoop to EmoLoop ($\Delta \textrm{Success}=0.119$, $\Delta \textrm{Sentiment}=0.30$). Such difference can be attributed to the lack of RL training in LLM-based systems. 


\subsection{Ablation Study}
\label{sec:analysis-further}

\begin{table}[h]
\centering
\scriptsize
\setlength\tabcolsep{3pt}
\begin{tabular}{llllcc}
\toprule
\bf System & Und & Gen & Man & \multicolumn{1}{c}{\bf Success} & \multicolumn{1}{c}{\bf Sentiment} \\ \midrule
SimpleLLAMA & - & - & - & 0.330 & 0.214 \\ 
 & + & - & - & 0.360 & 0.233 \\
 & - & + & - & 0.373 & 0.229 \\
EmoLLAMA & + & + & - & 0.342 & \textbf{0.250} \\ \midrule
SimpleLoop & - & - & S & 0.556 & 0.337 \\
 & + & - & S+E & 0.559 & 0.354 \\
 & - & + & S & 0.543 & 0.361 \\
EmoLoop & + & + & S+E & 0.531 & \textbf{0.405} \\ 

\bottomrule
\end{tabular}
\caption{Success and average user sentiment of systems from the interactive evaluation with langEmoUS. +/- means whether emotion is involved in the corresponding ToD stage: \textbf{Und}erstanding, \textbf{Man}agement, or \textbf{Gen}eration. For Management, ``-'' means the system is trained via SL,``S'' and ``E'' mean training via RL with success reward and emotion reward respectively. 
\label{tab:ablation}}
\end{table}

We ablate our emotional modular and end-to-end systems by incorporating emotion in different parts of the pipeline. Table \ref{tab:ablation} summarises their interactive performance with langEmoUS.


For both modular systems and end-to-end systems, incorporating emotion does not significantly change task success with the user simulator ($p>0.5$). The average user sentiment does improve slightly as emotion is introduced in understanding (plus management) and generation. Yet, the improvement from the non-emotional base system only becomes significant when emotion is added to all ToD stages. This highlights the importance of considering emotion in the whole ToD loop: it is necessary not only to understand user emotion but also to make use of it for dialogue management and respond with the appropriate conduct.~\footnote{See Appendix \ref{sec:ablation-emoloop-sl} for ablation study on EmoLoop with SL policy. A similar trend has been observed.}

\begin{figure}[ht]
    \centering
    \includegraphics[width=0.4\textwidth]{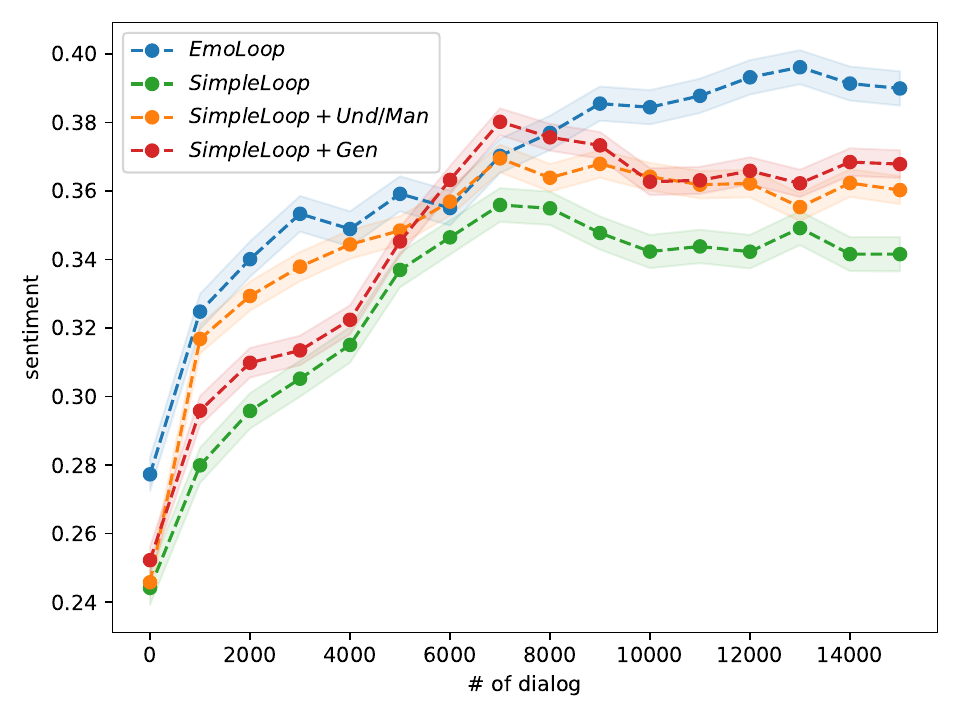}
    \caption{The average sentiment of langEmoUS during RL training of modular policy.}
    \label{fig:lang-sentiment}
\end{figure}

Figure \ref{fig:lang-sentiment} illustrates the change in the average sentiment of the simulated user during RL. At the beginning, average sentiments of modular systems fall in the similar range as SL-trained end-to-end systems, and are then further improved by RL. This highlights the importance of task success and emotion feedback signal for RL in ToD systems.


\section{Conclusion}
In this work, we incorporate emotion into the complete ToD processing loop, involving understanding, management, and generation. To achieve this, we first enrich the EmoWOZ dataset with system conduct labels to construct EmoWOZ 2.0. We then build modular and end-to-end ToD systems, as well as emotional user simulators with the newly collected dataset. We train the modular system policy via RL with the emotional user simulator and the end-to-end system via SL on EmoWOZ 2.0. Through interactive evaluation with both simulated and human users, we show that incorporating emotion into ToD systems can improve user's emotional experience as well as task success. 

There is still a long way to go from our work to the perfect emotional ToD system. Yet, we show our method as a promising avenue to achieve this ultimate goal. In our study, we directly translate user emotion labels into valence scores on a linear scale as a reward for RL. We believe that utilising the full set of user emotion labels for diverse reward would be a promising future direction.

We hope that with our work, we can motivate future research efforts to look at user experience beyond task success for ToDs and bring about insights to other task-oriented conversation settings. We would also like to highlight the opportunities in further improving LLM-based end-to-end ToD systems via RL, combining established approaches for policy training in modular systems and recent advancements in LLM research in other applications.

\section{Limitations}
One of the main limitations of modular ToD systems is the error accumulation in the pipeline for both modular and end-to-end systems. In modular systems, since each module is trained with a dataset associated with a limited ontology, the concepts that the system can understand and express are also limited. Although the DDPT policy, SetSUMBT DST, and many other models such as Trippy-R \citep{heck-etal-2022-robust} are built with the ability to handle out-of-domain requests, 
the generalisability and robustness of ToD systems are still challenges in the field that is yet to be solved.

All system modules have been trained in a supervised fashion on EmoWOZ 2.0. Therefore, the dataset contains limited dialogue situations and inherent bias. As seen in the dialogue examples in the appendix, the emotional responses are also limited. Yet, EmoWOZ 2.0 is the best resource we have at the moment. Data augmentation has been applied when training the NLG and the ERC model to mitigate the lack of diversity in the dataset. The RL training of the policy also allows the policy to explore more diverse dialogue trajectories. For the user simulator, considering data augmentation and more attributes of users, e.g. a more fine-grained user persona from chit-chat, would be a potential future direction to improve the diversity in simulated user behaviours.


Although LLMs can have better performance on each ToD modelling task and therefore could potentially serve as more powerful modules in EmoLoop, we did not move in this direction since their high computing resource requirement and slow inference speed would hinder their integration into our systems for interactive training and evaluation. Training modular system policy with langEmoUS for 15k dialogues on one Nvidia GeForce RTX 2080 Ti takes around 40 hours. The training time and memory required will be significantly increased if modular systems use LLM-based modules. On the other hand, while LLM-based end-to-end systems may provide a bypass since one LLM is sufficient, implementing RL training on such systems to further leverage task success and emotion signals from the user simulator is another computationally expensive challenge that are yet to solve.

Some of our generative system modules are based on pre-trained language models. Although we have not been reported any harmful generations in the human trail, there is still the possibility for unexpected behaviour when this system is deployed and tested on a very large scale.

For human evaluation, we conducted experiments on Amazon Mechanical Turk platform rather than deployed our systems in the production environment. The participants, despite coming from different countries, are  from covering all demographics.

\section{Ethics Statement}
Models, codes and datasets were used in accordance with their respective licenses, terms of use and intended use. The data that we used and generated does not contain any information that names or uniquely identifies individual people or offensive content. The model we used for generating augmented samples has implemented training objectives for enhanced safety (Appendix \ref{sec:nlg-appendix}). Systems we used for interaction with real users were very unlikely to generate offensive content as they were fine-tuned on large-scale training data to convey a limited scope of semantic concepts. No offensive content was reported by human users nor observed in post-hoc inspection.

For system conduct annotation, annotators were required to read and agree with our statement of consent for data use before the task. Annotators were paid fairly according to the local regulations of our research institute. We ensured swift communication with annotators so that their concerns were addressed as soon as possible. For poor-quality annotations, we still pay the annotators for their time but block them from our task to ensure data quality and collection efficiency. All annotations are anonymised.

The data annotation and interactive human trial, which involves decision making based on human emotions, have been approved by the ethics review board of the research institute. The proposed system learns how to manipulate human emotional state. Although the system is trained to elicit positive user emotion, this could still be of potential ethical concern and would require greater deliberation when deployed in real-life and more complex scenario.

\section{Acknowledgement}
S. Feng, N. Lubis, and M. Heck are supported by funding provided by the Alexander von Humboldt Foundation in the framework of the Sofja Kovalevskaja Award endowed by the Federal Ministry of Education and Research. H-C. Lin and C. van Niekerk are supported by the Ministry of Culture and Science of North Rhine-Westphalia within the framework of the Lamarr Fellow Network. C. Geishauser, B. Ruppik, and R. Vukovic are supported by funds from the European Research Council (ERC) provided under the Horizon 2020 research and innovation programme (Grant agreement No. STG2018804636). Computing resources were provided by Google Cloud.

\bibliography{anthology,custom}
\bibliographystyle{acl_natbib}

\appendix
\newpage
\onecolumn
\setcounter{table}{0}
\counterwithin{figure}{section}
\renewcommand{\thetable}{\Alph{section}\arabic{table}}

\section{EmoWOZ 2.0 Construction}
\subsection{Annotation Scheme Justification}
\label{sec:occ-justification}
Under the framework of the OCC emotion model and the definition of emotional empathy that the observer shares the emotional state of another person \citep{empathy}, we can derive the corresponding emotional response from the system. Considering the following user emotion and situation where:\\

The user is labelled as \emph{Fearful}, or feeling negative because of an event which has negative consequences on the user his or herself (as defined in EmoWOZ).\\

An empathetic operator would share the same feeling as the user (therefore also feeling negative). Yet, the feeling in the operator is elicited by an event which has negative consequences on the user (the other party). This feeling is defined as pity, or compassionate in the OCC model.

\subsection{Annotation Interface}
\label{sec:emowoz2.0-interface}
We adopted the same annotation set-up, annotator selection criteria, and quality assurance approaches as outlined by \citet{feng-etal-2022-emowoz}. Each utterance is annotated by three annotators, who were provided with the entire preceding dialogue history when annotating the current utterance. Annotators were English speakers. The final label was obtained from majority voting. When the agreement could not be reached, a fourth annotator was introduced. Overall, 54 crowd workers have contributed to our study. 

\begin{figure}[h]
    \centering
    \includegraphics[width=1\textwidth]{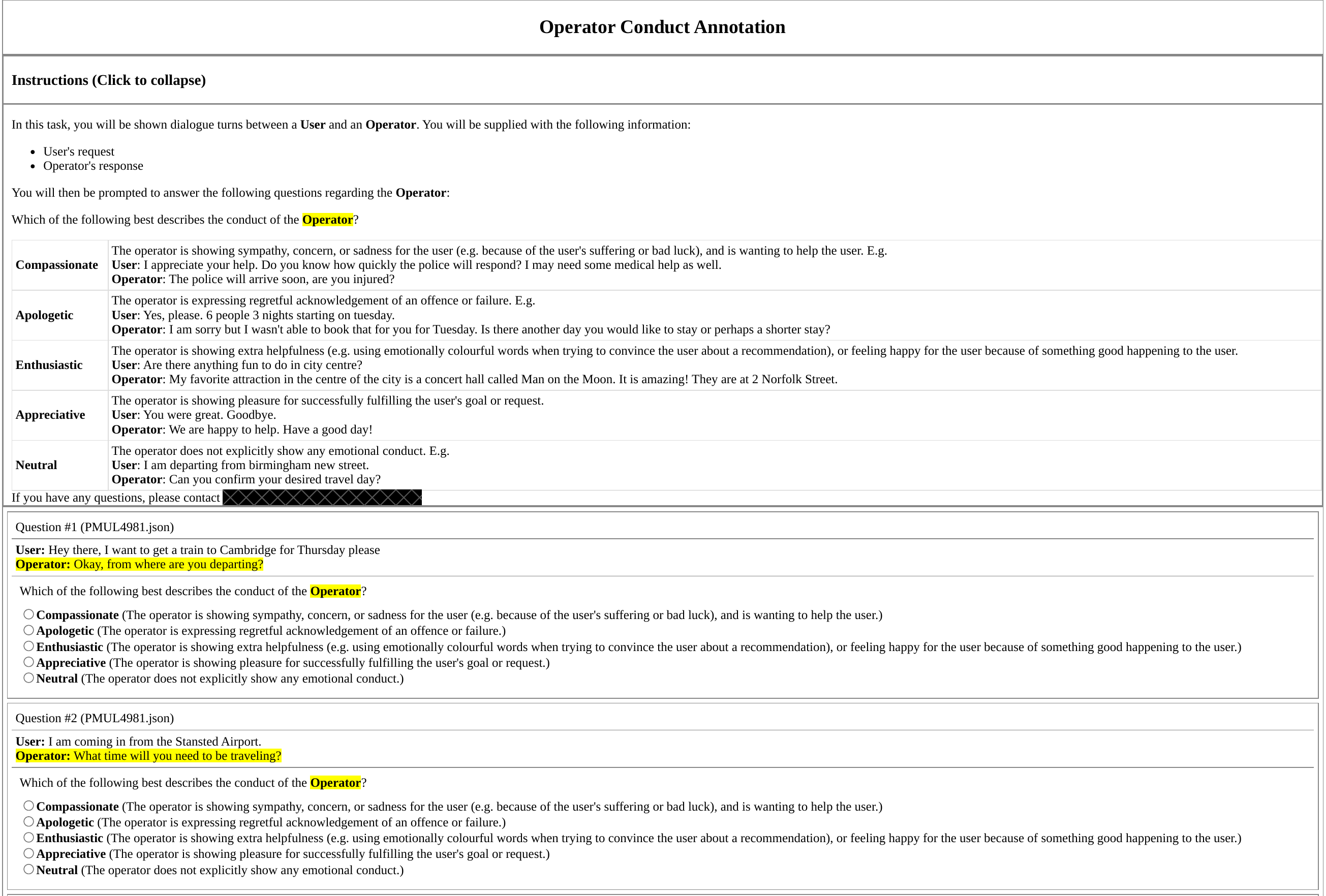}
    \caption{Web-interface for conduct annotation.}
    \label{fig:annotation-interface}
\end{figure}
\pagebreak
\subsection{Annotator Confusion Matrix}
\label{sec:emowoz2.0-confusion}

\begin{figure}[h]
    \centering
    \includegraphics[width=0.4\textwidth]{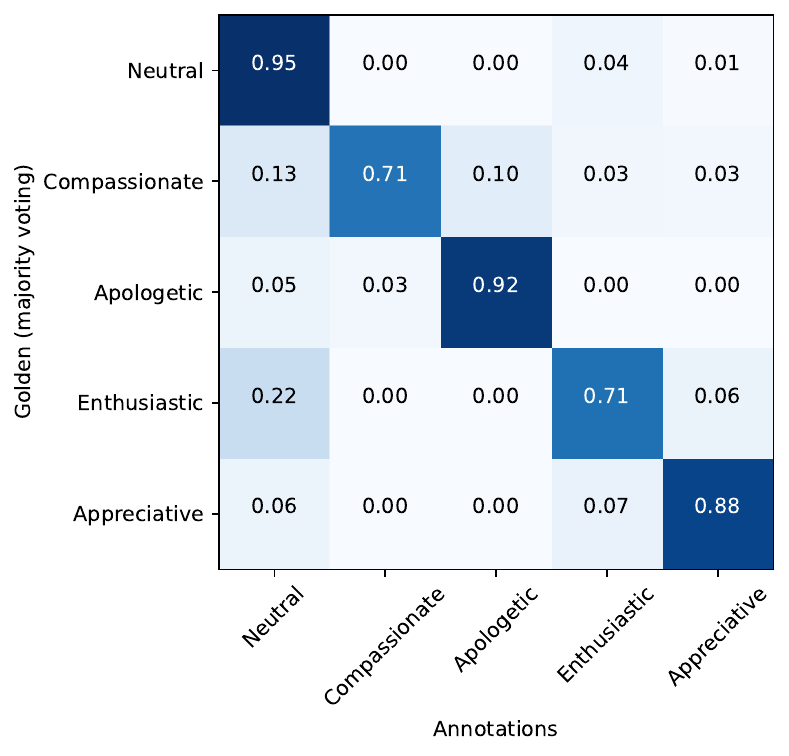}
    \caption{Annotator confusion matrix.}
    \label{fig:confusion-matrix}
\end{figure}


\subsection{System Conduct Distribution}
\label{sec:emowoz2.0-statistics}
\begin{table}[h]
\small
\centering
\begin{tabular}{lrr}
\toprule
\bf   Conduct           & \bf Count  & \bf Proportion \\ \midrule
Neutral       & 52,236 & 73.0\%     \\
Appreciative  & 9,763   & 13.6\%     \\
Enthusiastic  & 6,364   & 8.9\%      \\
Apologetic    & 3,049   & 4.3\%      \\
Compassionate & 112    & 0.2\%      \\ \bottomrule
\end{tabular} 
\caption{Conduct distribution in MultiWOZ. \label{tab:conduct-distribution}}
\end{table}

\section{User Simulator Implementation Details}
\label{sec:user-simulator-appendix}
\setcounter{table}{0}
Following the setting in \citet{emous}, the input and output of 
langEmoUS are represented as JSON-formatted strings, which are composed of tokens in natural language. We initialised our model based on the BART model~\cite{lewis2020bart} and fine-tuned it on our EmoWOZ 2.0 dataset. We optimised our model with Adam~\cite{DBLP:journals/corr/KingmaB14}, where the learning rate is $2e^{-5}$ for $5$ epochs.
As shown in Table~\ref{tab:emotional-user}, langEmoUS achieves state-of-the-art performance on user sentiment and emotion prediction. 

\begin{table}[h]
\centering
\small
\begin{tabular}{@{}lcc@{}}
\toprule
\bf Model         & \bf Sentiment      & \bf Emotion        \\ \midrule
SatActUtt~\citep{satactutt}      & 0.379          & -              \\
EmoUS~\citep{emous}         & 0.693          & 0.501          \\
langEmoUS     & \textbf{0.742} & \textbf{0.521} \\ \bottomrule
\end{tabular}
\caption{Performance for emotion and sentiment prediction of different models by measuring macro-F1 score.}
\label{tab:emotional-user}
\end{table}

\section{Natural Language Generator Implementation Details}
\label{sec:nlg-appendix}
\setcounter{table}{0}
\subsection{NLG Training}

\subsubsection{Training Configuration}

We trained SC-BART and SEC-BART on EmoWOZ 2.0.
We trained our model with Adam optimiser for standard cross entropy loss where the learning rate was set to $2e^{-5}$ for 5 epochs (with an early-stopping criterion based on the loss in the validation set) and a batch size of 16. During inference, we set the temperature to 0.9 and a beam number of 2 to promote some degree of diversity.

\subsubsection{Prompt Template} 
Our NLG models take the following input: previous user utterance $u_t$, dialogue semantic actions $a_t$, and conduct $e^{sys}_t$ (for SEC-BART only). The prompt template is shown as follows:

\paragraph{SEC-BART} Given the previous user request ``\{$u_t$\}'', the natural language realisation of dialogue action ``\{$a_t$\}'' with a/an ``\{$e^{sys}_t$\}'' conduct is

\paragraph{SC-BART} Given the previous user request ``\{$u_t$\}'', the natural language realisation of dialogue action ``\{$a_t$\}'' is

Given the prompt, the model predicted the probability distribution for a sequence of tokens. The output target is the corresponding ground-truth system response in EmoWOZ 2.0.

\subsubsection{Model Performance}
\begin{table}[h]
\centering
\small
\begin{tabular}{lcc}
\toprule
    \bf Model     & \bf BLEU $\uparrow$ & \bf SER $\downarrow$  \\ \midrule
SC-GPT \citep{peng-etal-2020-shot}  & 33.6                     & 4.8                     \\
T5NLG  \citep{zhu-etal-2023-convlab}  & 35.8                     & 3.7                     \\
SC-BART  & \textbf{35.9}                     & 3.9                     \\
SEC-BART & 34.9                     & \textbf{3.6} \\
\bottomrule
\end{tabular}
\caption{NLG Performance.\label{tab:nlg-performance}}
\end{table}

\subsection{Data Augmentation}
\subsubsection{Augmented Sample Collection}
Since the conduct distribution in EmoWOZ 2.0 is heavily imbalanced, we leveraged large language models for data augmentation. We selected system utterances with neutral conduct as the source to paraphrase for a target non-neutral conduct. We used LLaMA-2-13b-chat model \citep{touvron2023llama}. We used the following prompt:\\

Given the user request ``\{$u^{usr}_t$\}'' and the operator response action ``\{$a_t$\}'', please paraphrase the operator response ``\{$u^{sys}_{t,groundtruth}$\}'' in a more ``\{$e^{sys}_{t,target}$\}'' way? Please only give the answer, in less than $2\times len(u^{sys}_{t,groundtruth})$ tokens and enclosed with [RESP][/RESP].\\

We also experimented with ICL but the model tends to over-fit on the ICL samples. We therefore let it paraphrase in an zero-shot set-up to best explore its knowledge from pre-training for better diversity in the expression. 

\subsubsection{Augmented Sample Selection}
Since the model does not always follow the target conduct. For example, the large language model (LLM) would find some action-conduct combinations unreasonable. We therefore applied filtering on the LLM-generated samples.

\paragraph{Conduct Expressiveness} We trained an ensemble of 10 ContextBERT-ERToD models for conduct classification on EmoWOZ 2.0. The classifier reports an average weighted F1 score of 81.8\% without neutral. We then used majority voting from the classifier ensemble to correct the original target conduct when generating the sample.

\paragraph{Faithfulness to Semantic Action} We used the rule-based script in ConvLab-3 to evaluate NLG slot error rates in the paraphrased output based on the dialogue actions in the prompt. If there are slot errors in the output, we drop the sample.\\

Overall, we obtained 949 samples for \textit{Compassionate}, 900 for \textit{Apologetic}, 2274 for \textit{Enthusiastic}, and 490 for \textit{Appreciative}.

\section{Human Evaluation}
\label{sec:appendix-human-evaluation}
\setcounter{table}{0}

\subsection{Web Interface}
\label{sec:appendix-human-evaluation-web}
\begin{figure}[h]
    \centering
    \includegraphics[width=\textwidth]{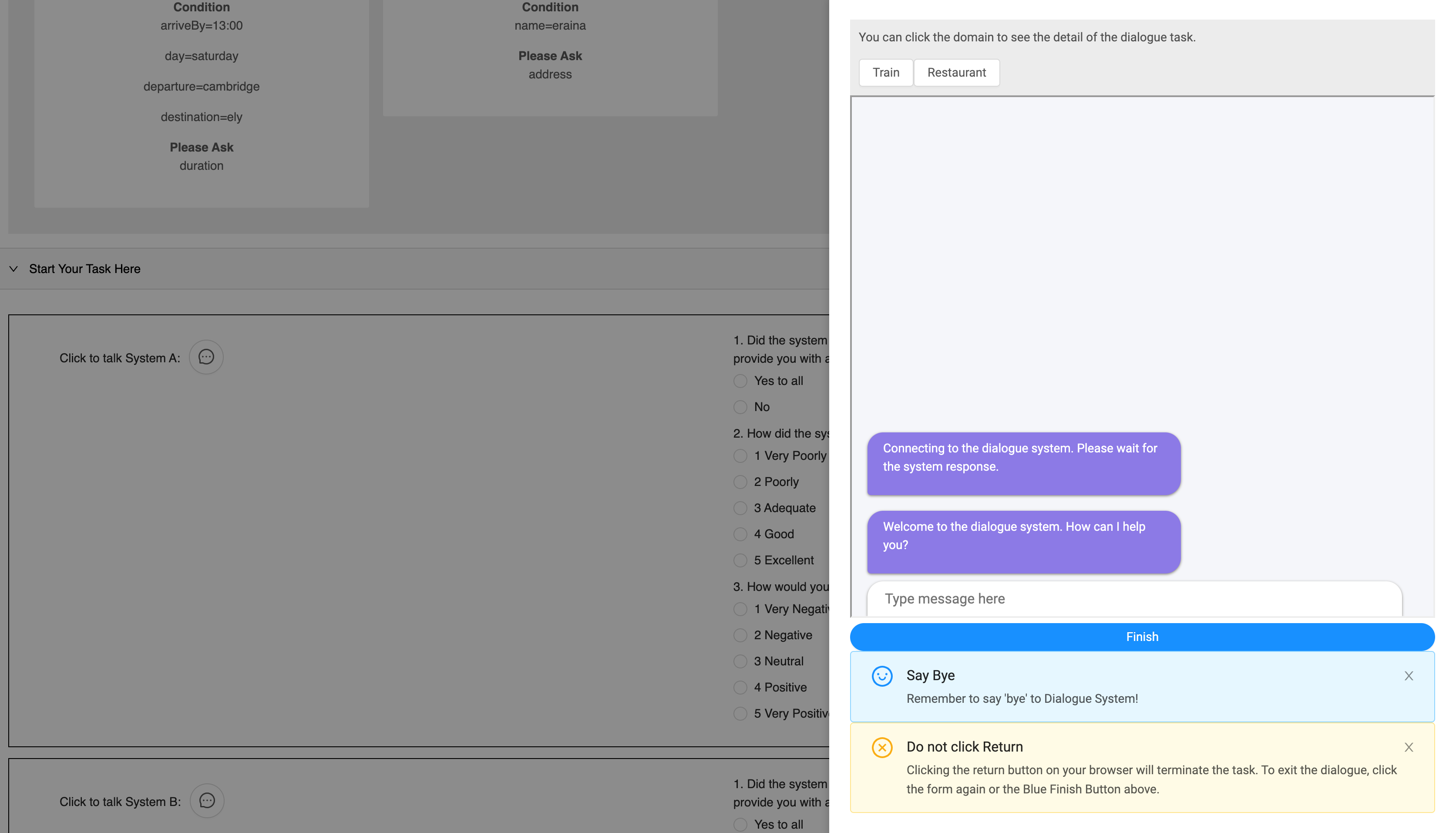}
    \caption{The web interface for human trial.}
    \label{fig:dialcrowd}
\end{figure}

\subsection{Survey Questions}
\label{sec:appendix-human-evaluation-survey}

\subsubsection{Question 1 - Task Success}
\paragraph{Question} Did the system find what you look for? Did it provide all the information that you need? If you ask for a booking, did it provide you with a reference number?
\paragraph{Multiple Choices} (A) Yes to all; (B) No.

\subsubsection{Question 2 - Sentiment Rating}
\paragraph{Question} How would you rate your sentiment after the conversation?
\paragraph{Multiple Choices} (A) Very Negative; (B) Negative; (C) Neutral; (D) Positive; (E) Very Positive.
\subsection{Dialogue Excerpts}
\label{sec:appendix-human-evaluation-example}

\begin{table}[H]
\small
\centering
\begin{tabular}{@{}llm{12cm}@{}}
\toprule
System & Role & Text \\ \midrule
EmoLoop & {\color[HTML]{FFB570}User} & Hello, I am looking for an attraction called byard art.  {\color[HTML]{3166FF}I am really excited about my trip to}\\
 & & {\color[HTML]{3166FF}Cambridge!!} \\
 & {\color[HTML]{97D077}System} & {\color[HTML]{3166FF}I'm sorry}, there is byard art in the south. The address is 14 kings parade. \\
 & {\color[HTML]{FFB570}User} & {\color[HTML]{3166FF}Why are you apologising?} \\
 & {\color[HTML]{97D077}System} & {\color[HTML]{3166FF}I apologize for the confusion.} If you need to contact them, please call 01223464646 \\ \midrule
SimpleLoop & {\color[HTML]{FFB570}User} & No, I just want to book a hotel for 3 people and 3 nights on friday and I need the address. \\
 & {\color[HTML]{97D077}System} & I can do that. What is the name of the hotel? \\
 & {\color[HTML]{FFB570}User} & Can you suggest one? \\
 & {\color[HTML]{97D077}System} & I have booked a taxi for you. It is a toyota and the contact number is 29929243243 \\
 & {\color[HTML]{FFB570}User} & {\color[HTML]{3166FF} Sorry, you are not helping at all} \\
 & {\color[HTML]{97D077}System} & I was able to book that for you. The reference number is 63484139414. What is your destination? \\ \bottomrule
\end{tabular}
\caption{Human trial dialogue excerpts from EmoLoop and SimpleLoop in a similar situation where both systems were not performing adequately at the beginning. EmoLoop responded with an improper conduct but realised the mistake and continued to provide information. SimpleLoop did not show any sign of realising the mistake.}
\end{table}

\begin{table}[H]
\small
\centering
\begin{tabular}{@{}llm{12cm}@{}}
\toprule
System & Role & Text \\ \midrule
EmoLoop & {\color[HTML]{FFB570}User} & I need the phone number of the hotel with name ashley hotel. \\
 & {\color[HTML]{97D077}System} & {\color[HTML]{3166FF}Sure thing!} It is located in the north and the postcode is cb41er and the phone number is \\
 & & 01223350059. \\
 & {\color[HTML]{FFB570}User} & {\color[HTML]{3166FF}Thanks a lot.} I also need a spanish restaurant with moderate price. \\
 & {\color[HTML]{97D077}System} & I would recommend la tasca. It is located at 14-16 bridge street. \\ \midrule
SimpleLoop & {\color[HTML]{FFB570}User} & {\color[HTML]{3166FF}Alright, I am also interested} in a museum in the city centre. \\
 & {\color[HTML]{97D077}System} & I recommend broughton house gallery. It is a museum. Their phone number is 01223314960. \\
 & {\color[HTML]{FFB570}User} & {\color[HTML]{3166FF}Nice,} could you also give me the address and postcode? \\
 & {\color[HTML]{97D077}System} & The postcode is cb11ln and the phone number is 01223314960. \\ \bottomrule
\end{tabular}
\caption{Human trial dialogue excerpts from EmoLoop and SimpleLoop in a similar situation where the user expressed excitement and appreciation. Both systems performed adequately in terms of completing the user's goal. EmoLoop responded in an enthusiastic way whereas SimpleLoop did not respond with affective conduct.}
\end{table}

\begin{table}[H]
\small
\centering
\begin{tabular}{@{}llm{12cm}@{}}
\toprule
System & Role & Text \\ \midrule
EmoLoop & {\color[HTML]{FFB570}User} & I need some information. \\
 & {\color[HTML]{97D077}System} & What kind of information do you need? \\
 & {\color[HTML]{FFB570}User} & Which department there? \\
 & {\color[HTML]{97D077}System} & {\color[HTML]{3166FF}I am sorry, I don't understand. What do you mean?} \\
 & {\color[HTML]{FFB570}User} & I need information about haematology. \\
 & {\color[HTML]{97D077}System} & The Addenbrookes Hospital is located at Hills Rd, Cambridge , postcode cb20qq. The phone \\
 & & number is 01223216297. \\  \midrule
SimpleLoop & {\color[HTML]{FFB570}User} & I need information about haematology day unit. \\
 & {\color[HTML]{97D077}System} & The pizza hut cherry hilton Hospital is located at Hills Rd, Cambridge , cb20qq. The phone \\
 & & number is 01223216297. \\
 & {\color[HTML]{FFB570}User} & Department details? \\
 & {\color[HTML]{97D077}System} & There are 10 trains that go to london kings cross on sunday. The TR7423 leaves at 05:00 and \\
 & & arrives at 05:51. Would you like me to book that for 6 people? \\ \bottomrule
\end{tabular}
\caption{Human trial dialogue excerpts from EmoLLAMA and SimpleLLAMA in a similar situation where the user asked for hospital information. EmoLLAMA, although not understanding the user's vague request, attempted to apologise and clarify the request with the user. The task was successfully completed. SimpleLLAMA carried on with errors and fails the task. (The wrong values, ``pizza hut'' and ``cherry hilton'', in the SimpleLLAMA response were due to wrong domains in the dialogue action prediction, which led to corresponding name slots in the response. These name slots were filled during lexicalisation based on actions, resulting in an obviously irrational output.)}
\end{table}

\begin{table}[H]
\small
\centering
\begin{tabular}{@{}llm{12cm}@{}}
\toprule
System & Role & Text \\ \midrule
EmoLoop & {\color[HTML]{FFB570}User} & Recommend a taxi which arrives by 4:15 pm from the fez club to tandoori place. \\
 & {\color[HTML]{97D077}System} & {\color[HTML]{3166FF}I'd be happy to help with your request,} first I'll need to know where you'll be leaving from, \\
 & & where you're going, and what time you'd like to arrive. \\
 & {\color[HTML]{FFB570}User} & I will be leaving from fez club and going to tandoori palace arriving by 4:15 pm. \\
 & {\color[HTML]{97D077}System} & Booking was successful. The car type is bmw and the contact number is 19287599612. \\  \midrule
SimpleLoop & {\color[HTML]{FFB570}User} & I want to book a taxi for 16:15 from cherry hinton water play to ely train station. \\
 & {\color[HTML]{97D077}System} & What is your destination? \\
 & {\color[HTML]{FFB570}User} & Ely train station. \\
 & {\color[HTML]{97D077}System} & Booking completed. Booked car type: ford, Contact number: 46563428252. \\ \bottomrule
\end{tabular}
\caption{Human trial dialogue excerpts from EmoLLAMA and SimpleLLAMA in a similar situation where both systems failed to capture all information provided in the user request. EmoLLAMA at first missed the information provided by the user but replied in a compassionate way. The user repeated and then the system provides the correct information. Likewise, SimpleLLAMA missed the destination in the first turn. After the user repeated, the system completed the task for the user. Yet, there is no affective interaction between the user and SimpleLLAMA.}
\end{table}

\section{Further Analysis}
\label{sec:ablation-appendix}
\setcounter{table}{0}
\subsection{Ablation Study for EmoLoop with Supervised Training Only}
\label{sec:ablation-emoloop-sl}
\begin{table}[H]
\centering
\setlength\tabcolsep{3pt}
\begin{tabular}{llllcc}
\toprule
\bf System & Und & Gen & Man & \multicolumn{1}{c}{\bf Success} & \multicolumn{1}{c}{\bf Sentiment} \\ \midrule
SimpleLoop-SL & - & - & - & 0.512 & 0.244 \\
 & + & - & - & 0.494 & 0.246 \\
 & - & + & - & 0.493 & 0.249 \\
EmoLoop-SL & + & + & - & 0.516 & 0.273 \\
\bottomrule
\end{tabular}
\caption{Success and average user sentiment of our system variants from the interactive evaluation with langEmoUS. +/- means whether the emotion is involved in the corresponding ToD stage: \textbf{Und}erstanding, \textbf{Man}agement, or \textbf{Gen}eration. All systems are trained via SL.
\label{tab:ablation-sl}}
\end{table}

\subsection{Impact of Training Set-ups on System Conduct}
\begin{figure*}[h]
     \centering
     \begin{subfigure}[b]{0.32\textwidth}
         \centering
         \includegraphics[width=\textwidth]{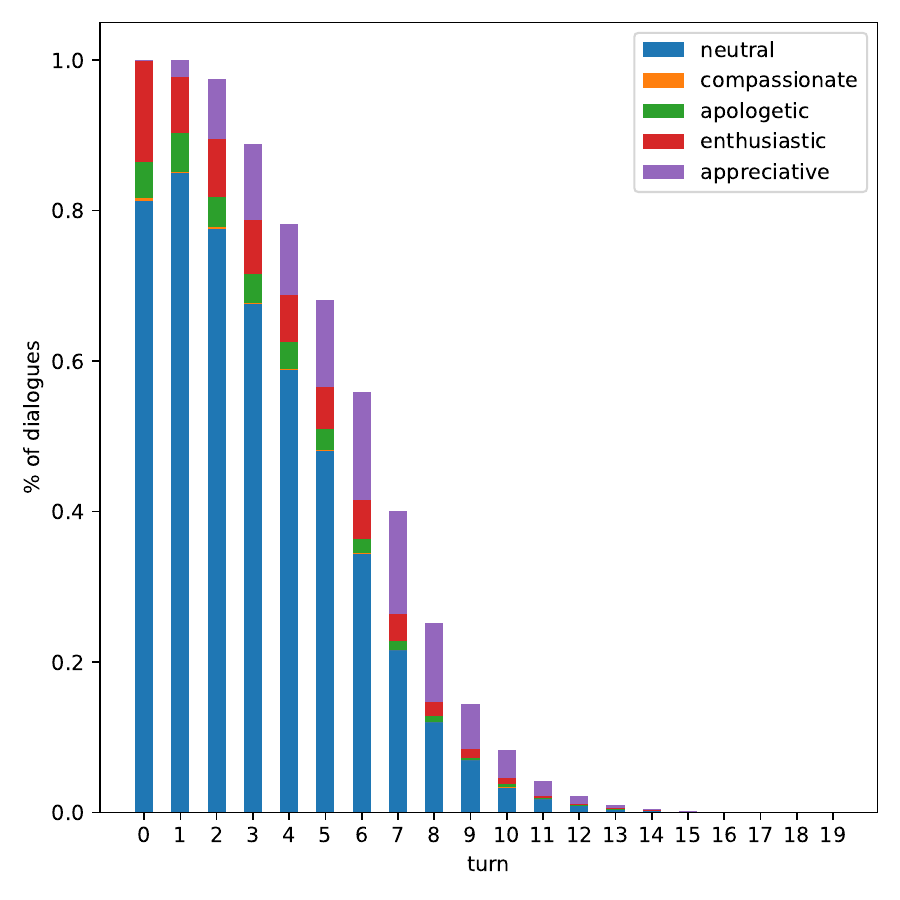}
         \caption{Dataset Distribution}
         \label{fig:conduct-dataset}
     \end{subfigure}
     \hfill
     \begin{subfigure}[b]{0.32\textwidth}
         \centering
         \includegraphics[width=\textwidth]{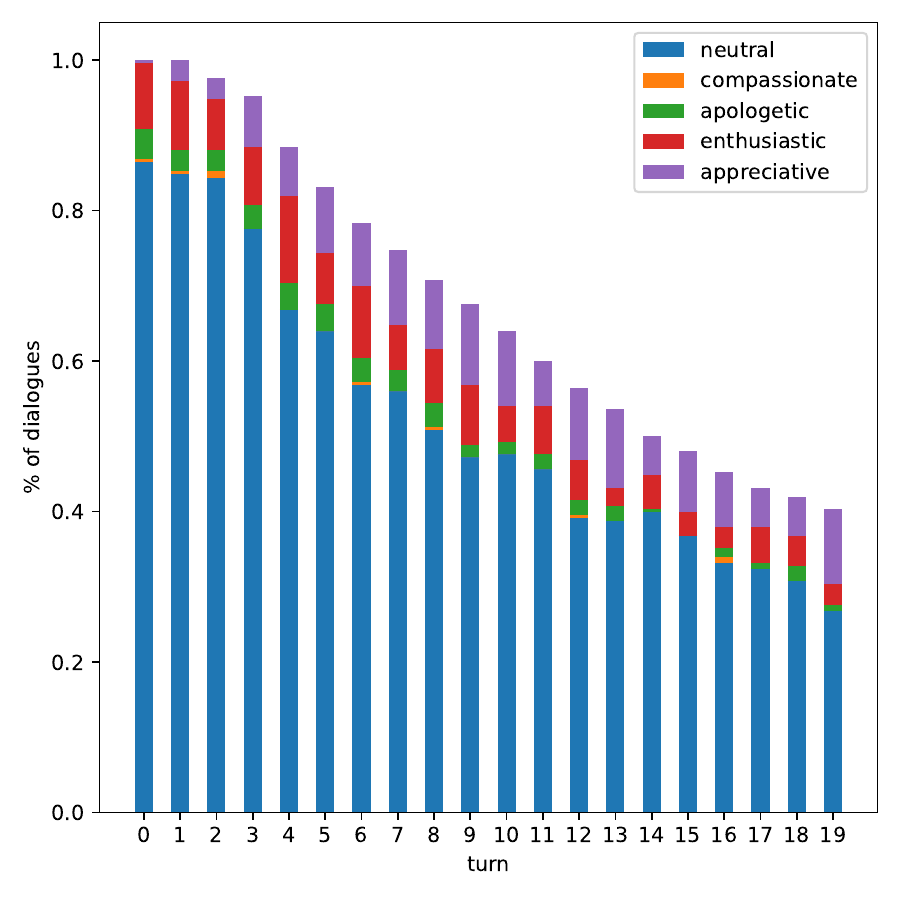}
         \caption{Supervised Pre-training}
         \label{fig:conduct-sl}
     \end{subfigure}
     \hfill
     \begin{subfigure}[b]{0.32\textwidth}
         \centering
         \includegraphics[width=\textwidth]{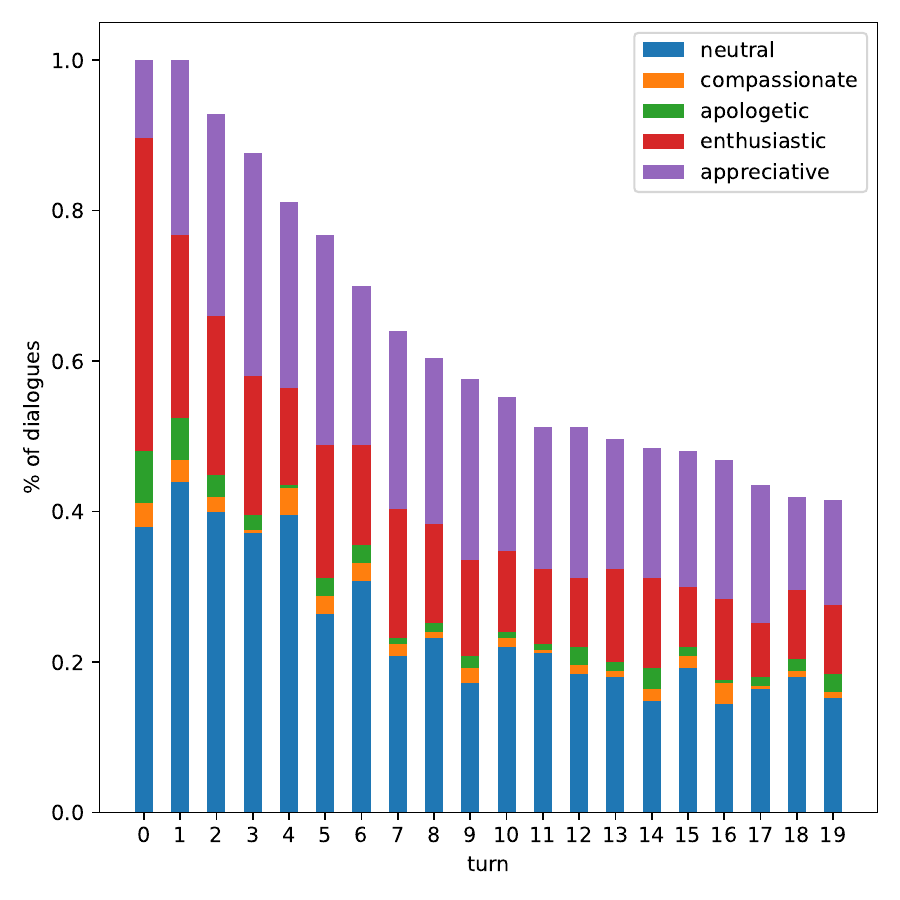}
         \caption{RL on Language Level}
         \label{fig:conduct-rl-language}
     \end{subfigure}
        \caption{Distributions of system conduct for different turn positions at different stages of policy training.}
        \label{fig:conduct-distribution}
\end{figure*}

We investigate how the EmoLoop's affective behaviour is shaped in different stages of training. Figure \ref{fig:conduct-distribution} shows the distribution of system conduct at different dialogue turns in EmoWOZ 2.0, and policy output during interaction with langEmoUS after supervised pre-training and language-level RL. Comparing Figure \ref{fig:conduct-dataset} and Figure \ref{fig:conduct-sl} suggests that the policy imitates the affective behaviour of operators in the corpus. 

After RL, the policy is more inclined to express \textit{enthusiastic} and \textit{appreciative} while expressing \textit{compassionate} and \textit{apologetic} less frequently. This illustrates the affective strategy of the policy to elicit more positive emotions in the simulated user.

\end{document}